\def\year{2020}\relax
\documentclass[letterpaper]{article} % DO NOT CHANGE THIS
\usepackage{subcaption}
\usepackage{aaai20}  % DO NOT CHANGE THIS
\usepackage{times}  % DO NOT CHANGE THIS
\usepackage{helvet} % DO NOT CHANGE THIS
\usepackage{courier}  % DO NOT CHANGE THIS
\usepackage[hyphens]{url}  % DO NOT CHANGE THIS
\usepackage{graphicx} % DO NOT CHANGE THIS
\usepackage{graphicx}  % DO NOT CHANGE THIS
\usepackage{xspace}
\newcommand{\astar}{A$^*$\xspace}
\newcommand{\wastar}{WA$^*$\xspace}
\newcommand{\focalsipp}{FocalSIPP\xspace}
\newcommand{\wdsipp}{WSIPP$_d$\xspace}
\newcommand{\wrsipp}{WSIPP$_r$\xspace}
\newcommand{\wsipp}{WSIPP\xspace}

\usepackage[pscoord]{eso-pic}% The zero point of the coordinate systemis the lower left corner of the page (the default).

\title{Revisiting Bounded-Suboptimal Safe Interval Path Planning\thanks{This is a pre-print of the paper accepted to ICAPS 2020.}}
\author{
\Large \textbf{Konstantin Yakovlev,\textsuperscript{\rm 1, \rm 2} Anton Andreychuk,\textsuperscript{\rm 1, \rm 3} Roni Stern\textsuperscript{\rm 4, \rm 5}}\\  %All authors must be in the same font size and format. Use \Large and \textbf to achieve this result when breaking a line
\textsuperscript{\rm 1}Federal Research Center for Computer Science and Control of Russian Academy of Sciences\\
\textsuperscript{\rm 2}National Research University Higher School of Economics\\
\textsuperscript{\rm 3}Peoples’ Friendship University of Russia (RUDN University)\\
\textsuperscript{\rm 4}Ben-Gurion University of the Negev, \textsuperscript{\rm 5}Palo Alto Research Center (PARC)\\
yakovlev@isa.ru, andreychuk@mail.com, sternron@post.bgu.ac.il % email address must be in roman text type, not monospace or sans serif
}

\begin{document}

%\placetextbox{0.5}{0.95}{\fbox{\texttt{This is a pre-print of the paper accepted to ICAPS 2020}}}%

\maketitle

\begin{abstract}
Safe-interval path planning (SIPP) is a powerful algorithm for finding a path in the presence of dynamic obstacles. SIPP returns provably optimal solutions. However, in many practical applications of SIPP such as path planning for robots, one would like to trade-off optimality for shorter planning time. In this paper we explore different ways to build a bounded-suboptimal SIPP and discuss their pros and cons. We compare the different bounded-suboptimal versions of SIPP experimentally. While there is no universal winner, the results provide insights into when each method should be used. 
\end{abstract}

\section{Introduction}
Finding a shortest path in a graph is a classical problem in computer science with numerous applications, including robot motion planning, digital entertainment, and logistics. \astar~\cite{hart1968formal} and Dijsktra's algorithm~\cite{dijkstra1959note} are well-known methods for solving this kind of task. Finding a path becomes more challenging in the presence of dynamic obstacles that move through the environment, blocking vertices or prohibiting moving between some of them at specific time ranges. A solution in such a scenario is a \emph{plan} that consists a sequence of \emph{actions}, where an action is either to move from one vertex to an adjacent one, or to wait in it for some time. Application of textbook \astar or Dijkstra's algorithm in this setting is not straightforward. First, the set of actions in a vertex depends on the current time, due to the dynamic obstacles. Second, there are potentially infinite wait actions, depending on how much time one would like to wait. 

To address this problem, the \emph{Safe Interval Path Planning} (SIPP) algorithm was introduced~\cite{phillips2011}. SIPP computes for vertices in the graph a set of \emph{safe intervals} in which it is possible to occupy them without colliding with the dynamic obstacles. Then, it runs an \astar search in a different graph in which each vertex represents a pair of vertex in the original graph and a safe time interval. SIPP is complete and returns optimal solutions. It has been successfully applied in a range of domains, including robot motion planning and multi-agent path finding~\cite{araki2017multi,andreychuk2019multi,cohen2019optimal}. In such applications, a common requirement is to tradeoff solution optimality in order to obtain a solution faster. To control this tradeoff, we explore \emph{bounded-suboptimal} versions of SIPP. A bounded-suboptimal algorithm accepts a parameter $w\geq 1$ and returns a solution whose cost is at most $w$ times the cost of an optimal solution. Since SIPP is based on \astar, it is natural to apply the same frameworks used for creating a bounded-suboptimal \astar. However, the SIPP search space has certain properties that prevent a straightforward application of frameworks such as \wastar~\cite{pohl1970heuristic}. To address this, Narayanan et al.~\shortcite{narayanan2012anytime} proposed a bounded-suboptimal SIPP implementation. 

In this work we revisit the assumptions of this algorithm and propose two alternative bounded-suboptimal SIPP algorithms. One based on the \wastar framework but allowing node re-expansions (see description later), and another based on the \emph{focal search} framework~\cite{pearl1982studies}. We analyze these algorithms and compare them experimentally. The results show that each algorithm has its strengths and weaknesses, and the choice of which algorithm should be use depends mainly on the value of $w$. 

\section{Problem Statement}

Consider a mobile agent that navigates in an environment represented by a weighted graph. The vertices of the graph correspond to the locations the agent may occupy, and the edges correspond to allowed transitions. When at a vertex, the agent can either \textit{wait} for an arbitrary amount of time, or \textit{move} to an adjacent vertex along a graph edge. For the sake of simplicity we neglect the inertial effects and assume that agent moves with constant speed such that the duration of a move action equals the weight of the corresponding edge. 

A \emph{plan} $\pi(s, g)$ is a sequence of  consecutive actions that move the agent from a start vertex $s$ to goal vertex $g$. The cost of a plan is the sum of durations of its constituent actions. There exist dynamic obstacles that move in the environment and block certain vertices and edges for pre-defined time intervals, preventing the agent from occupying or moving through them. A plan $\pi(s,g)$ is called \emph{valid} if it avoids colliding with all dynamic obstacles. The \emph{path planning with dynamic obstacles problem} is the problem of finding a valid plan for a given start and goal locations. 

An \emph{optimal} solution to this problem is a lowest-cost valid plan from start to goal. The \emph{suboptimality} of a solution is the ratio between its cost and the cost of an optimal solution. In this work we are interested in finding solutions whose suboptimality is bounded by a given scalar $w \geq 1$. E.g. by setting $w$ to $1.1$ we aim to find a solution whose cost is no more than 10\% over the cost of an optimal solution. 

\section{Background}

\astar is a heuristic search algorithm for finding a path in a state space represented as a graph. It maintains two lists of states, OPEN and CLOSED. OPEN contains all states that were generated but not expanded and CLOSED contains all previously expanded states. Initially, CLOSED is empty and OPEN contains only the initial state. Every state $n$ is associated with two values: $g(n)$, which is the cost of the lowest-cost path found so far from the initial state to $n$, and $h(n)$, which is a heuristic estimate of the cost of the lowest-cost path from $n$ to a goal. In every iteration the state with the minimal $f(n)=g(n)+h(n)$ is popped out of OPEN and \emph{expanded}. It means generating the \emph{successors}, which are state's neighbors in the state space, and inserting them into OPEN. \astar halts when it expands a goal state. Note that if a state $n$ generates a state $n'$ that was already generated, then we must check if its $g$ value can be updated by considering reaching it via $n$. If this happens and $n'$ is no longer in OPEN, then it must be re-inserted into OPEN. Consequently, a node may be \emph{re-expanded} multiple times. In fact, in some extreme cases the number of expanded states can be exponential in the size of the state space~\cite{martelli1977complexity}. 

\astar has several attractive properties that are relevant for this paper. First, if $h(n)$ is \emph{admissible}, that is, it always outputs a lower bound on the value it estimates, then \astar guarantees finding an optimal solution (if one exists). Second, if heuristic is \emph{consistent}, then  \astar will never re-expand a state. A heuristic is consistent if for every pair of states $n$ and $n'$ such that $n$ generates $n'$ it holds that $h(n)-h(n')\leq c(n,n')$ where $c(n,n')$ is the cost of the edge from $n$ to $n'$. 

SIPP is a modification of \astar for the path planning with dynamic obstacles problem. The core idea of SIPP is to group consecutive time moments into \textit{time intervals} and to associate every vertex with one or more safe intervals. A safe interval for a vertex is ``a contiguous period of time ... during which there is no collision and it is in collision one timestep prior and one timestep after the period''~\cite{phillips2011}. 
SIPP performs an \astar search over a state space in which a state is a tuple $(v, [t_i, t_j])$, where $v$ is a vertex in the underlying graph and $[t_i,t_j]$ is a \emph{safe interval} for $v$. Note that states with different non-overlapping time intervals but the same vertex might exist in the search space. 

$g(n)$ in SIPP is the earliest time an agent can reach $v$ in the designated safe interval $[t_i, t_j]$. SIPP uses $g(n)$ to compute the set of successors for $n$. I.e., when a state $n=(v, [t_i, t_j])$ generates a state $n'=(v', [t'_i, t'_j])$, SIPP first tries to commit a move $v \rightarrow v'$ at time equal to $g(n)$, that is without any waiting at $n$. If such a transition results in a collision with a dynamic obstacle, then SIPP augments the move with a wait action of minimal duration. That is, the agent is planned to wait at $v$ no longer than it is needed to avoid the collision. Therefore, $g(n')$ is set to be $g(n)+dur(min_{wait})+c(v,v')$, where $c(v,v')$ is the cost of the edge $(v,v')$ and $dur(min_{wait})$ is computed taking dynamic obstacles into account.\footnote{The method to compute $dur(min_{wait})$ is domain dependent.} 

$h(n)$ in SIPP estimates the lowest-cost plan to move the agent from $n$ to a goal state. SIPP requires $h(n)$ to be both admissible and consistent. Beyond these changes, SIPP uses regular \astar: it extract from OPEN the state with the lowest $g(n)+h(n)$ and it halts when a goal has been expanded. 

SIPP shares several of the desirable properties of \astar. It guarantees finding a solution if one exists (otherwise, it reports \textit{failure}) and this solution is optimal. Both completeness and optimality rely on the following property: when SIPP expands a state $n=(v, [t_i, t_j])$, then $g(n)$ is the earliest possible arrival time to $v$ in $[t_i, t_j]$ (Theorem 2 in~\cite{phillips2011}). This property guarantees that set of successors for every generated state is maximal (Theorem 1 in~\cite{phillips2011}). Consequently, expanding a state with the goal vertex means the lowest cost plan to it has been found and all relevant states have been generated.  

\section{Bounded-Suboptimal SIPP}
One of the well-known ways to make \astar bounded-suboptimal is to use an inflated heuristic function during the search. \wastar~\cite{pohl1970heuristic} is a prominent example of this approach. \wastar is similar to \astar except that it chooses for the expansion a node $n$ that minimizes $g(n)+w \cdot h(m)$, where $w \geq 1$ is the desired suboptimality bound. When a goal state is expanded, \wastar is guaranteed to have found a solution whose suboptimality is at most $w$.

Unlike \astar, when \wastar expands a state $n$ it may be that $g(n)$ is \emph{not} the lowest cost path to $n$. Thus, a state may expanded more than once. Nevertheless, if $h(n)$ is consistent, \wastar is guaranteed to find a solution with the desired suboptimality bound even without re-expanding a single state~\cite{likhachev2004ara}. 

\subsection{Weighted SIPP (\wsipp)}
Let Weighted SIPP (\wsipp) be SIPP that uses an inflated heuristic like \wastar. \wsipp is not guaranteed to find a bounded-suboptimal solution, or in fact, any solution, if it does not re-expand states. For example, see Figure \ref{fig:RunningExample}.\footnote{The rectangles above the vertices are explained later.} A dynamic obstacle moves from $A$ to $B$, arriving at $B$ at time 10 and staying there forever. Thus, $[0, 10]$ is the only safe interval for $B$. The safe interval for all states except $A$ and $B$ is $[-\infty,\infty]$. Let $w=2$. After the first expansion states $n_D=(D,[-\infty,\infty])$ and $n_E=(E,[-\infty,\infty])$ are generated. $n_E$ is chosen for expansion as $f(n_E)=4+7\cdot2 < f(n_D)=3+8\cdot2$. Next, node $n_C=(C,[-\infty,\infty])$ is generated and its $g$-value is set: $g(n_C)=8$. Then $n_C$ is expanded but it has no successors, as one can not reach $B$ within the safe interval $[0, 10]$ from $C$ when $g(n_C)=8$ (as $8+3>10$). To find a solution, we must expand $n_D$ and then re-expand $n_C$ to update $g(n_C)$ to 6. 

\subsubsection{Weighted SIPP with Duplicate States}
The creators of SIPP identified this problem and proposed the following. When the initial state is expanded, we create two copies of every state it generates. An \emph{optimal copy}, which is prioritized in OPEN according to $w \cdot (g+h)$ and a \emph{suboptimal copy}, which is prioritized in OPEN according to $g+w \cdot h$. Throughout the search, whenever an optimal copy is expanded, we again generate two copies of every state it generates. We refer to this algorithm as \emph{\wsipp with Duplicate States} (\wdsipp). \wdsipp preserves the desirable property of \wastar -- it guarantees returning a bounded-suboptimal solution while avoiding re-expansions. 
That is, every copy of a state is expanded at most once. 

\subsubsection{Weighted SIPP with Re-expansions (\wrsipp)}
Interestingly, the creators of SIPP did not explore the possibility of allowing unlimited re-expansions instead of duplicating each state. That is, if a state in CLOSED is generated with a lower $g$ value, then it is re-inserted to OPEN. Thus, every state will eventually be re-expanded with the minimal $g$ value that guarantees finding a solution. We call this algorithm \emph{\wsipp with Re-expansions} (\wrsipp). 

\wrsipp is complete and is guaranteed to find a bounded-suboptimal solution, but the number of re-expansions performed by \wrsipp can be exponential~\cite{martelli1977complexity}. Nevertheless, in many domains the number of re-expansions is manageable~\cite{sepetnitsky2016repair}. 

\begin{figure}[t]
    \centering
    \includegraphics[width=0.95\columnwidth]{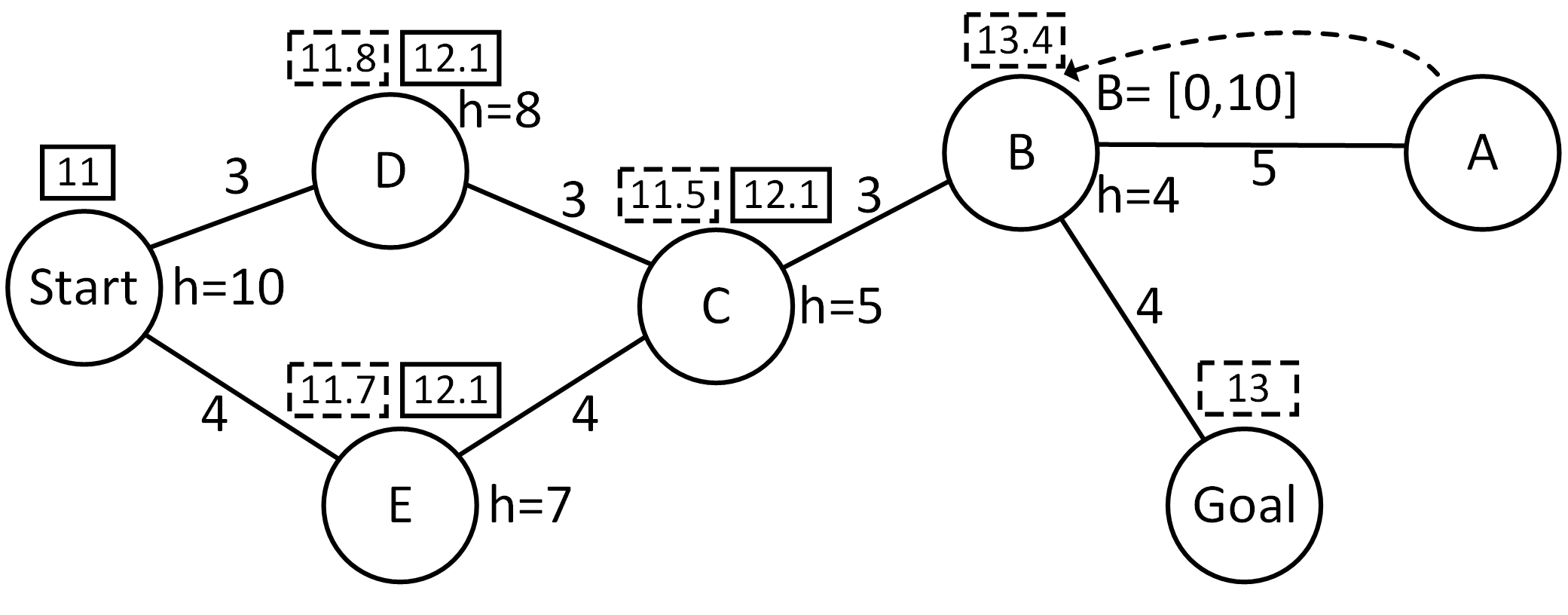}
    \caption{A running example for \wsipp.}
    \label{fig:RunningExample}
\end{figure}

In fact, \wrsipp may expand fewer states than \wdsipp. Consider Fig.~\ref{fig:RunningExample}. Rectangles over the vertices show the states expanded by \wdsipp with $w=1.1$, where the optimal and sub-optimal copies of each state are marked by a solid and dashed line. The values inside the rectangles are $f$ values. After the initial expansion, \wdsipp expands sub-optimal $n_E$ with $f=11.7$ and generates sub-optimal $n_C$ with $f=13.5$. Then sup-optimal $n_D$ ($f=11.8$) is expanded. This adjusts the $f$ value of sub-optimal $n_C$ to $11.5$. Next, \wdsipp expands the latter and generates sub-optimal $n_B$ with $f=13.4$. Then optimal $n_D$ and optimal $n_E$ are consequently expanded as their $f$ values (both equal to $12.1$) are lower than $13.4$. As a result of these expansions optimal $n_C$ is generated and its $f$ value is set to $12.1$. It is expanded and optimal $n_B$ with $f=14.3$ is generated. Then \wdsipp switches back to sup-optimal $n_B$ (as $13.4 < 14.3$) and expands it generating sub-optimal copy of $Goal$ with $f=13$. Finally, the latter is expanded and the search terminates. Total number of expansions is $9$. In the same setting \wrsipp performs only $6$ expansions: $Start$, $n_E$, $n_D$, $n_C$, $n_B$, $Goal$, with no re-expansions at all. 

\subsection{Focal SIPP}
Focal Search~\cite{pearl1982studies} is a framework for bounded-suboptimal search that maintains a sublist of OPEN, called FOCAL. FOCAL is the set of every state $n$ for which $g(n)+h(n)\leq w\cdot f_{min}$, where $f_{min}$ is the smallest $f$ value over all states in OPEN. In every iteration, a state $n$ is chosen from FOCAL that minimizes a secondary heuristic $h_F(n)$. The latter does not have to be consistent or even admissible. A well-known secondary heuristics is the number of hops-to-the-goal~\cite{wilt2014speedy}, which ignores edge costs. 
Focal Search terminates when a goal state is in the FOCAL or when OPEN is empty.

\begin{figure*}[t]
\centering

\includegraphics[width=\linewidth]{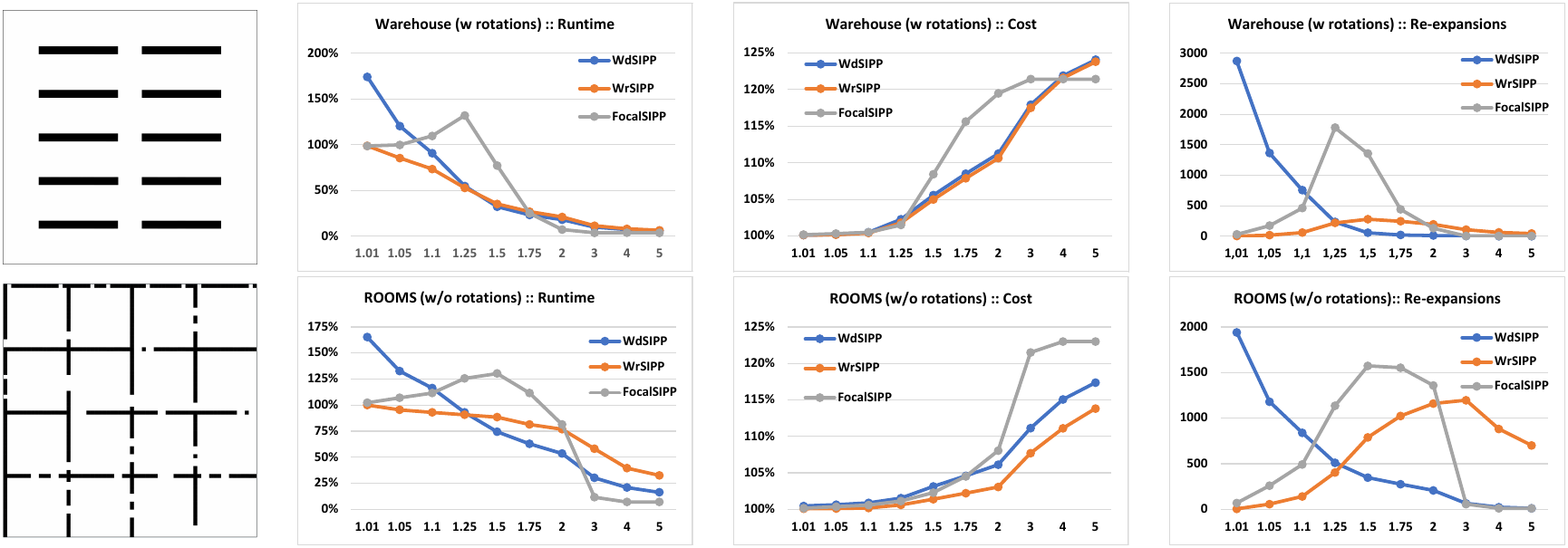}

\caption{From left to right: map of the environment, runtime, solution cost, number of re-expansions. Top row: Warehouse w rotations. 
Bottom row: Rooms w/o rotations.}
\label{fig:results}
\end{figure*}

We use the name \emph{Focal SIPP} (\focalsipp) to refer to the SIPP version that uses Focal Search instead of \astar and allows unlimited re-expansions. \focalsipp is complete and guarantees finding a bounded-suboptimal solution. However, due to re-expansions its runtime may be exponential in the number of states in the search space, just like \wrsipp.

\subsection{Discussion}
\wdsipp, \wrsipp, and \focalsipp are guaranteed to return a bounded-suboptimal solution. However, the cost of the solution they return may differ, since it can be any value between the optimal cost and $w$ times that cost. Perhaps, more interesting, though, is the expected runtime for each algorithm for a given value of $w$. 

Consider setting $w$ to be very close to one. In this case, \wdsipp is expected to perform poorly, since two copies of every generated state are introduced and both are likely to have nearly the same priority in OPEN, since $w \cdot (g+h)$ and $g+w\cdot h$ are very close if $w$ is close to one, especially in the beginning of the search. In contrast, \wrsipp is expected to perform well, as it is almost equivalent to \astar, which is known to expand the minimal number of states~\cite{dechter1985generalized}. On the other hand, consider the behavior of \focalsipp for very large values of $w$. In such a case almost all states in OPEN will be in FOCAL and \focalsipp basically performs a greedy best-first search towards a goal, which is specifically designed to reach a goal state quickly. Thus, we expect in these cases that \focalsipp will work well. In the experimental results below we confirmed these expectations. 

\section{Experimental Results} 
We evaluated the considered algorithms on a range of different grid maps including empty $64\times 64$ map, $64\times 64$ map containing 10 rectangular obstacles that resemble a warehouse (\textit{Warehouse}), $64\times 64$ map composed of square rooms connected by the passages (\textit{Rooms}), $257 \times 256$ game map.\footnote{The Rooms map and the game map (den520) were taken from the movingai repository~\cite{sturtevant2012}. Our code can be found at https://github.com/PathPlanning/SuboptimalSIPP.} Grid connectivity varied from 8- to 32-connected. Each map was populated with 250 dynamic obstacles that move between random cells. 100 different pairs of start and goal locations for an agent were chosen on each map randomly. Two action models were considered -- one that assumes only translations (denoted ``w/o rotations'') and one that also considers rotations (denoted ``w rotations''), i.e., if moving from one cell to the other requires aligning the heading, the agent has to rotate and it takes time. Moving speed was 1 cell per 1 time unit, rotation speed was $\pi / 2$ per time unit. We used the Euclidean distance, scaled properly with the agent's speed, as a heuristic. The secondary heuristic for \focalsipp, $h_F(n)$, was set to be the shortest path from $n$ to the goal ignoring all dynamic obstacles and edge costs, which we computed offline. The sub-optimality bound $w$ varied from $1.01$ to $5$. 

In each run we measured the algorithm's runtime and solution cost, relative to SIPP, and the number of re-expansions. While \wdsipp prohibits re-expansions, it eventually generates two copies of the same state. If both of them got expanded, we count this as a re-expansion.

Fig.~\ref{fig:results} shows the results of our experiments on two representative setups: \emph{Warehouse w/o rotations} (32-connected) and \emph{Rooms w rotations} (32-connected). We choose to present results for only these two setups due to space limitations. These specific maps represent two possible ``types'' of worlds. Warehouse is a relatively open environment populated with the isolated obstacles (similar to city maps, empty maps etc.), Rooms is a corridor-like environment with a large number of narrow passages (similar to mazes, indoor maps, etc.). The heuristic we used is relatively accurate for Warehouse but it is not accurate for Rooms, allowing us to show the impact of heuristic accuracy. Moreover, in the \textit{w rotations} model, the heuristic is even less accurate.

In general, we observed similar trends across all the considered domains and action models. These trends, supported by the Fig.~\ref{fig:results}, are the following: (1) \wrsipp is better for small values of $w$ that are close to 1, (2) \focalsipp is better or the same on higher $w$, and (3) \wdsipp is better or the same for mid-range $w$. For mid-range sub-optimality bounds the results also show a notable ``spike'' in the number of re-expansions and runtime for \focalsipp. We hypothesize that this occurs because the bound is large enough to cause finding suboptimal paths to generated nodes but not large enough so that bounded-suboptimal solutions can be found without re-expanding these nodes.

\begin{figure}
    \centering
    \includegraphics[width=0.75\columnwidth]{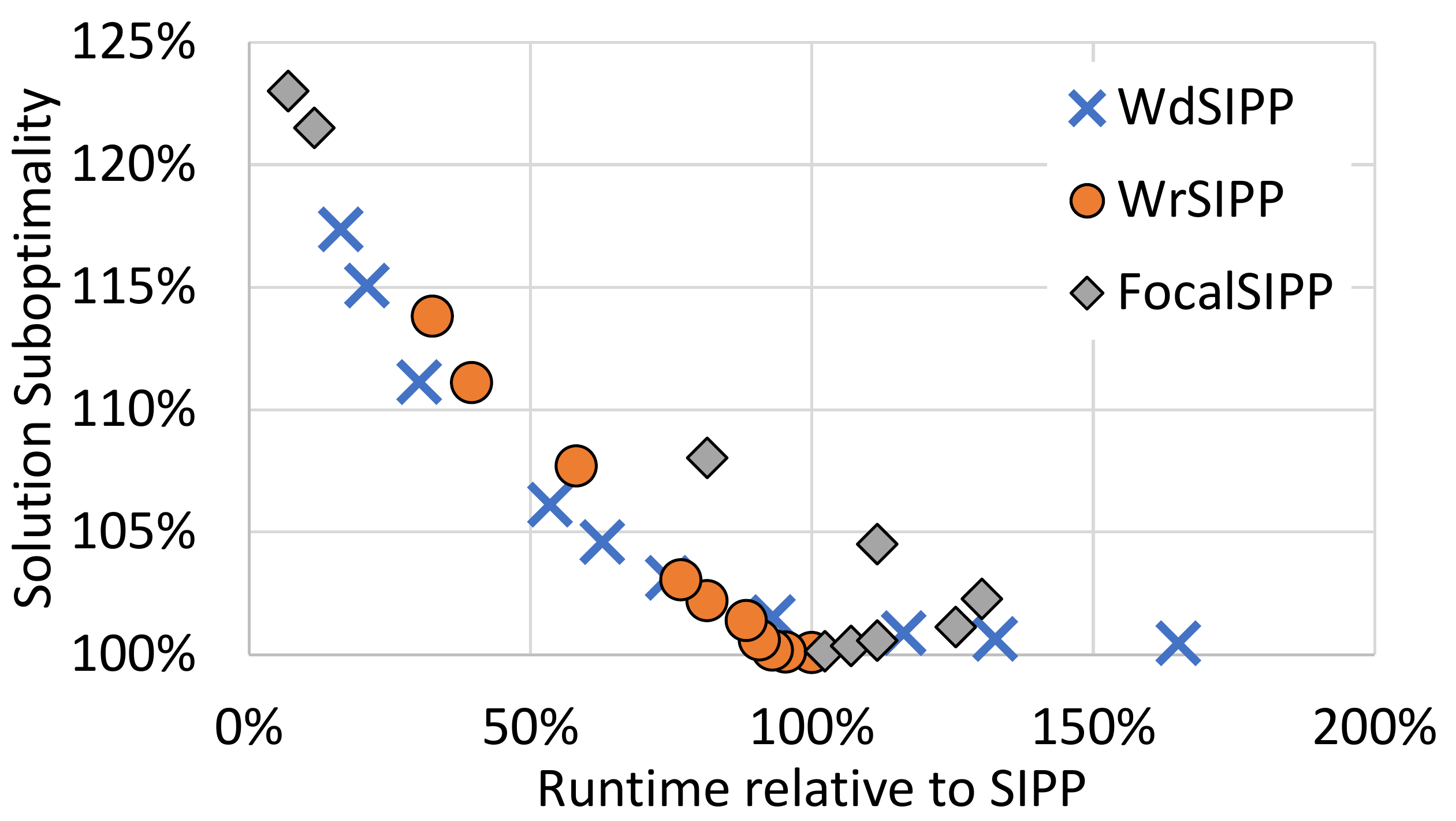}
    \caption{Solution suboptimality vs. relative runtime, for Room maps w/o rotations.}
    \label{fig:cost-vs-runtime}
\end{figure}

Figure~\ref{fig:cost-vs-runtime} shows the tradeoff between  solution cost and runtime obtained by \wdsipp, \wrsipp, and \focalsipp.  Each data point represents the results of an algorithm with a specific $w$ value, where the $x$ value is the average runtime and the $y$ value is the solution suboptimality. The results again show the same trends as above. \focalsipp allows the fastest solution when solution cost is the worst. \wrsipp is preferable if one wants very close to optimal solution, while \wdsipp is more suitable for mid-range suboptimality.

\section{Summary}
We explored three bounded-suboptimal versions of SIPP and analyzed their pros and cons. Experimental evaluation on different settings show that the previously proposed bounded-suboptimal SIPP -- \wdsipp~-- is frequently outperformed by the other algorithms, e.g. by \wrsipp, which runs \wastar but allows re-expanding. An appealing direction of future research is to explore more sophisticated techniques for bounded-suboptimal SIPP, as well as explore its applications in an anytime planning and multi-agent path finding.

\section{Acknowledgments}
This work was partially funded by RFBR (project 18-37-20032). Anton Andreychuk is supported by the ``RUDN University Program 5-100''. Roni Stern is supported by ISF grant \#210/17.

\bibliographystyle{aaai}
\bibliography{library}

\begin{thebibliography}{}

\bibitem[\protect\citeauthoryear{Andreychuk \bgroup et al\mbox.\egroup
  }{2019}]{andreychuk2019multi}
Andreychuk, A.; Yakovlev, K.; Atzmon, D.; and Stern, R.
\newblock 2019.
\newblock Multi-agent pathfinding with continuous time.
\newblock In {\em International Joint Conference on Artificial Intelligence
  (IJCAI)},  39--45.

\bibitem[\protect\citeauthoryear{Araki \bgroup et al\mbox.\egroup
  }{2017}]{araki2017multi}
Araki, B.; Strang, J.; Pohorecky, S.; Qiu, C.; Naegeli, T.; and Rus, D.
\newblock 2017.
\newblock Multi-robot path planning for a swarm of robots that can both fly and
  drive.
\newblock In {\em IEEE International Conference on Robotics and Automation
  (ICRA)},  5575--5582.

\bibitem[\protect\citeauthoryear{Cohen \bgroup et al\mbox.\egroup
  }{2019}]{cohen2019optimal}
Cohen, L.; Uras, T.; Kumar, T.~S.; and Koenig, S.
\newblock 2019.
\newblock Optimal and bounded-suboptimal multi-agent motion planning.
\newblock In {\em Symposium on Combinatorial Search}.

\bibitem[\protect\citeauthoryear{Dechter and
  Pearl}{1985}]{dechter1985generalized}
Dechter, R., and Pearl, J.
\newblock 1985.
\newblock Generalized best-first search strategies and the optimality of a.
\newblock {\em Journal of the ACM (JACM)} 32(3):505--536.

\bibitem[\protect\citeauthoryear{Dijkstra}{1959}]{dijkstra1959note}
Dijkstra, E.~W.
\newblock 1959.
\newblock A note on two problems in connexion with graphs.
\newblock {\em Numerische mathematik} 1(1):269--271.

\bibitem[\protect\citeauthoryear{Hart, Nilsson, and
  Raphael}{1968}]{hart1968formal}
Hart, P.~E.; Nilsson, N.~J.; and Raphael, B.
\newblock 1968.
\newblock A formal basis for the heuristic determination of minimum cost paths.
\newblock {\em IEEE transactions on Systems Science and Cybernetics}
  4(2):100--107.

\bibitem[\protect\citeauthoryear{Likhachev, Gordon, and
  Thrun}{2004}]{likhachev2004ara}
Likhachev, M.; Gordon, G.~J.; and Thrun, S.
\newblock 2004.
\newblock {ARA*}: Anytime {A*} with provable bounds on sub-optimality.
\newblock In {\em Advances in neural information processing systems (NIPS)},
  767--774.

\bibitem[\protect\citeauthoryear{Martelli}{1977}]{martelli1977complexity}
Martelli, A.
\newblock 1977.
\newblock On the complexity of admissible search algorithms.
\newblock {\em Artificial Intelligence} 8(1):1--13.

\bibitem[\protect\citeauthoryear{Narayanan, Phillips, and
  Likhachev}{2012}]{narayanan2012anytime}
Narayanan, V.; Phillips, M.; and Likhachev, M.
\newblock 2012.
\newblock Anytime safe interval path planning for dynamic environments.
\newblock In {\em IEEE/RSJ International Conference on Intelligent Robots and
  Systems},  4708--4715.

\bibitem[\protect\citeauthoryear{Pearl and Kim}{1982}]{pearl1982studies}
Pearl, J., and Kim, J.~H.
\newblock 1982.
\newblock Studies in semi-admissible heuristics.
\newblock {\em IEEE transactions on pattern analysis and machine intelligence}
  (4):392--399.

\bibitem[\protect\citeauthoryear{Phillips and Likhachev}{2011}]{phillips2011}
Phillips, M., and Likhachev, M.
\newblock 2011.
\newblock {SIPP}: Safe interval path planning for dynamic environments.
\newblock In {\em Proceedings of The 2011 IEEE International Conference on
  Robotics and Automation ({ICRA} 2011)},  5628--5635.

\bibitem[\protect\citeauthoryear{Pohl}{1970}]{pohl1970heuristic}
Pohl, I.
\newblock 1970.
\newblock Heuristic search viewed as path finding in a graph.
\newblock {\em Artificial intelligence} 1(3-4):193--204.

\bibitem[\protect\citeauthoryear{Sepetnitsky, Felner, and
  Stern}{2016}]{sepetnitsky2016repair}
Sepetnitsky, V.; Felner, A.; and Stern, R.
\newblock 2016.
\newblock Repair policies for not reopening nodes in different search settings.
\newblock In {\em Symposium on Combinatorial Search ({SOCS})},  81--88.

\bibitem[\protect\citeauthoryear{Sturtevant}{2012}]{sturtevant2012}
Sturtevant, N.~R.
\newblock 2012.
\newblock Benchmarks for grid-based pathfinding.
\newblock {\em IEEE Transactions on Computational Intelligence and AI in Games}
  4(2):144--148.

\bibitem[\protect\citeauthoryear{Wilt and Ruml}{2014}]{wilt2014speedy}
Wilt, C.~M., and Ruml, W.
\newblock 2014.
\newblock Speedy versus greedy search.
\newblock In {\em Symposium on Combinatorial Search (SoCS)}.

\end{thebibliography}

\end{document}